\documentclass{article}

\usepackage{arxiv}

\usepackage[utf8]{inputenc} % allow utf-8 input
\usepackage[T1]{fontenc}    % use 8-bit T1 fonts
\usepackage{hyperref}       % hyperlinks
\usepackage{url}            % simple URL typesetting
\usepackage{amsfonts}       % blackboard math symbols
\usepackage{nicefrac}       % compact symbols for 1/2, etc.
\usepackage{microtype}      % microtypography
\usepackage{multirow}
\usepackage{tablefootnote}
\usepackage{footnote}
\usepackage{graphicx}
\usepackage{afterpage}
\usepackage[comma,authoryear,round]{natbib}
\usepackage{fullpage}
\usepackage{caption}
\usepackage{needspace}
\usepackage{times}
\usepackage{fancyhdr,graphicx,amsmath,amssymb}

\title{Data Anomaly Detection for Structural Health Monitoring of Bridges using Shapelet Transform}

\author{
  Monica Arul\\
  NatHaz Modeling Laboratory\\
  Department of Civil Engineering\\
  University of Notre Dame\\
  Notre Dame, IN 46556 \\
  \texttt{maruljay@nd.edu} \\
   \And
 Ahsan Kareem \\
  NatHaz Modeling Laboratory\\
  Department of Civil Engineering\\
 University of Notre Dame \\
  Notre Dame, IN 46556 \\
  \texttt{kareem@nd.edu} \\
}

\begin{document}
\maketitle

\begin{abstract}
With the wider availability of sensor technology through easily affordable sensor devices, a number of Structural Health Monitoring (SHM) systems are deployed to monitor vital civil infrastructure. The continuous monitoring provides valuable information about the health of structure that can help in providing a decision support system for retrofits and other structural modifications. However, when the sensors are exposed to harsh environmental conditions, the data measured by the SHM systems tend to be affected by multiple anomalies caused by faulty or broken sensors. Given a deluge of high-dimensional data collected continuously over time, research into using machine learning methods to detect anomalies are a topic of great interest to the SHM community. This paper contributes to this effort by proposing the use of a relatively new time series representation named “Shapelet Transform” in combination with a Random Forest classifier to autonomously identify anomalies in SHM data. The shapelet transform is a unique time series representation that is solely based on the shape of the time series data. In consideration of the individual characteristics unique to every anomaly, the application of this transform yields a new shape-based feature representation that can be combined with any standard machine learning algorithm to detect anomalous data with no manual intervention. For the present study, the anomaly detection framework consists of three steps: identifying unique shapes from anomalous data, using these shapes to transform the SHM data into a local-shape space and training machine learning algorithm on this transformed data to identify anomalies. The efficacy of this method is demonstrated by the identification of anomalies in acceleration data from a SHM system installed on a long-span bridge in China. The results show that multiple data anomalies in SHM data can be automatically detected with high accuracy using the proposed method.
\end{abstract}

% keywords can be removed

\section{Introduction}
As the demands on the infrastructure continue to increase, research into structural health monitoring (SHM) has grown in importance throughout the world. The widespread application of sophisticated SHM systems in civil infrastructure produces a large volume of data. However, the harsh environmental conditions of civil structures cause the data measured by SHM systems to be affected by multiple anomalies caused by faulty or broken sensors. These anomalies pose a significant barrier for assessing the true structural performance and severely affects the automatic warning system for damage or accidents. The identification and removal of data anomalies due to environmental variations is thus an important preprocessing step in a successful warning system.
Several model-based methods have been developed in the past few decades for data anomaly detection in SHM data \citep{abdelghani2004sensor,thiyagarajan2017predictive,wan2018bayesian,wang2019modeling}. In these methods a number of statistical models are initially constructed to predict the measurements. Using appropriate thresholds, measurements that show significant differences between predicted and measured values are identified and treated as anomalies.

Faced with a massive amount of data due the continuous monitoring of structures, researchers have recently resorted to advanced approaches such as data mining and machine learning techniques for anomaly detection. \cite{bao2019computer} proposed a computer vision and deep learning–based data anomaly detection method in which the raw time series measurements are first transformed into image vectors which are then fed into the Deep Neural Networks (DNN) to train them to identify various anomalies from SHM data.\cite{fu2019sensor} used a similarity test based on power spectral density to detect anomalies and then trained an artificial neural network to identify the different types of sensor anomalies. \cite{tang2019convolutional} proposed the use of a Convolutional Neural Network (CNN) for anomaly detection that learned from multiple graphical information. The visualizations of the time series measurements in time and frequency domain are fed to the neural networks which then learned the characteristics of each of the anomalies during training. The trained network  is then used to identify and classify various anomalies. \cite{mao2020toward} used Generative Adversarial Networks (GAN) in combination with autoencoders to identify anomalies. The raw time series from the SHM system are first transformed into Gramian Angular Field (GAF) images which are then used to train the GAN and autoencoders to identify anomalies.

This paper contributes to this effort by proposing the use of a relatively new time series representation named “Shapelet Transform” in combination with Random Forest classifiers for anomaly detection in SHM data. The shapelet transform is a unique time series representation technique that is solely based on the shape of the time series. The raw measurements of every sensor anomaly has a unique time series shape. The shapelet transform utilizes this feature to easily capture these distinct shapes and the Random Forest classifier uses these shapes to identify and classify the different anomalous data patterns from a large SHM system database. 

Analysis methods based on the global attributes of time series are unintuitive and reduce comprehensibility. By examining local-shape-based features, it is ensured that these small discriminatory shapes are not averaged out but rather used to distinguish the time series, exactly as they are under intuitive visual inspection. The primary advantage of shapelets over the above mentioned competing methods is the interpretability and insight it offers. Shape-based approaches are intuitive, visually meaningful and offer immediate insight into the problem domain that goes beyond their use in accurate detection. Thus the method is a “white-box” machine learning model involving understandable and easily visualizable features that makes the entire process transparent and completely open to inspection. This, in turn, increases the interpretability of the model compared to the other state-of-the-art black-box machine learning models, like neural networks, and helps domain practitioners gain better insights from their data. In terms of applicability, shapelets have been utilized in a wide variety of domains including motion-capture \citep{ye2009time,ye2011time,lines2012shapelet,hartmann2010gesture}, spectrographs \citep{ye2009time,ye2011time}, tornado prediction \citep{mcgovern2011identifying}, detection of natural hazards\citep{arul2020applications}, medical and health informatics \citep{ghalwash2013extraction,xing2011extracting,xing2012early} among others. In the present study, the efficacy of this method is demonstrated by the identification of anomalies in SHM data obtained from a long-span bridge in China.

The article is organized as follows. A general overview of the shapelet transform is provided in section 2.  A brief description about the SHM data used for this study is given in the section 3. The methodology for the proposed anomaly detection process is elaborated in section 4. In this section, the different stages involved in shapelet transform are explained in great detail along with illustrative examples. The section also explains the step-by-step procedure for detection of anomalous patterns in SHM data obtained from the long-span bridge. Finally, a comprehensive summary of the anomalies detected using shapelet transform is provided in sections 5 and 6.

\section{Overview of shapelet transform}

 Consider time series 1 and 2 generated as a result of an event as shown in Fig. 1. Both the time series have long stretches of aperiodic waveforms. However, a local shape appears for a short duration in the time series that differs substantially from the rest of the time series. These localized shapes are called shapelets. These discriminatory shapes which are phase independent serve as a powerful feature for identifying anomalous patterns or classifying events from a large database containing continuous records.
 \begin{figure}[htbp]
  \centering
  \captionsetup{justification=centering}
  \includegraphics[scale=0.8]{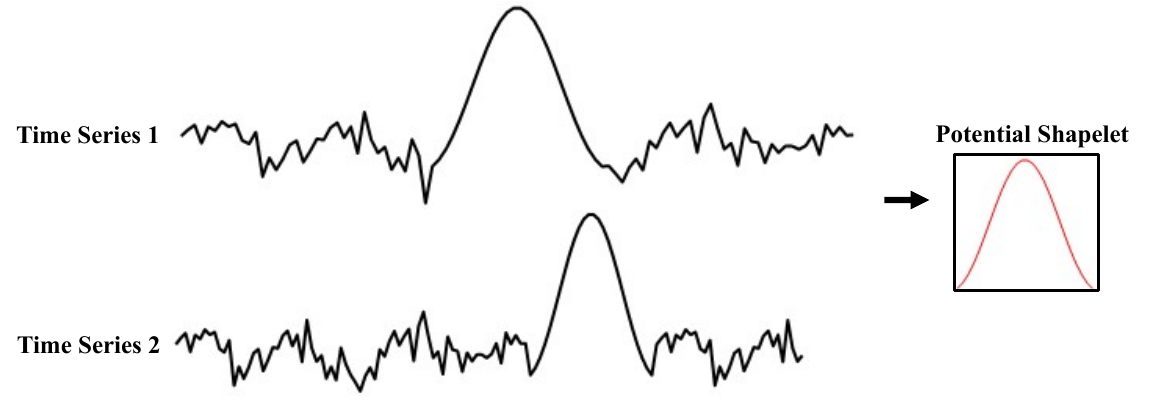}
  \caption{Time series shapelets}
  \label{fig:fig1}
\end{figure}
 Time series shapelets stem from the desire to reify human’s innate capacity to visualize the shape of data and identify almost instantly similarities and differences between patterns. Shapelets help computers perform this complex task by identifying the local or global similarity of shape that can offer an intuitively comprehensible way of understanding continuous time series. The shapelets, once discovered, can then be used to transform data into a local-shape space where each feature is the distance between a shapelet and a time series \citep{lines2012shapelet}. The result of this transform is that the new representation can be applied to any standard machine learning algorithm, to identify anomalous patterns. Shapelet transform has five major stages: generation of shapelet candidates, distance calculation between a shapelet and a time series, assessment of the quality of shapelets, discovery of shapelets, and data transformation.Each of these stages will be elaborated in detail in the following sections.

\section{Data description}

In this paper, SHM dataset from a long-span cable-stayed bridge in China is used. The main span of the bridge is 1088 m, two side spans are 300m each and it consists of two towers that are 306 m high. The structural health monitoring system of the bridge consists of 38 sensors, whose locations are illustrated in Fig. 2. The sensors include accelerometer, anemometer, strain gauge, global positioning system (GPS), and thermometer. For the present case, one-month (2012-01-01 – 2012-01-31) of acceleration data for all 38 sensors from the SHM system is considered for anomaly detection. The sampling frequency of the accelerometers is 20Hz. The continuous raw measurements are broken down into 1-hour segments and 744 time series measurements for each sensor is obtained for a one-month period resulting in a total of 744*38 datasets.The characteristics of the normal data and the six classes of anomalies found in the dataset is described in Table 1. Examples for each data pattern is shown in Fig. 3.  The normal time series measurement is labelled as 1 and the other six data anomaly patterns are labelled from 2 – 7. From Table 1, it can be seen that nearly 52\% of the data are anomalous. The “trend” is the major anomalous pattern constituting of 20\% of the dataset followed by “missing” and “square” each accounting for around 10\%. On the other hand, the “outlier” pattern accounts for only 1.9\% of the dataset followed by “drift” that constitutes of 2.4\% of the data.

\begin{figure}[htbp]
  \centering
  \captionsetup{justification=centering}
  \includegraphics[scale=0.9]{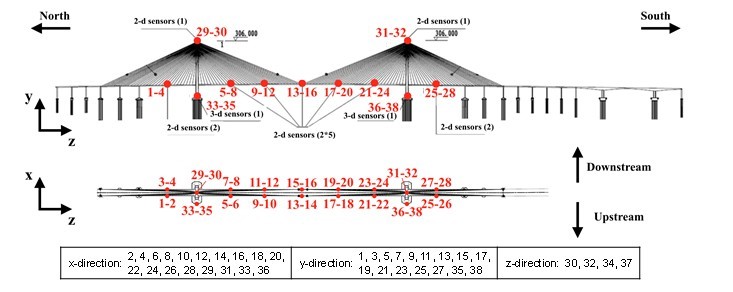}
  \caption{The bridge and the placement of accelerometers on the deck and towers}
  \label{fig:fig2}
\end{figure}

\begin{table}[htbp]
\begin{center}
\captionof{table}{Description of anomalous data patterns} \label{foobar}
\includegraphics[scale=0.9]{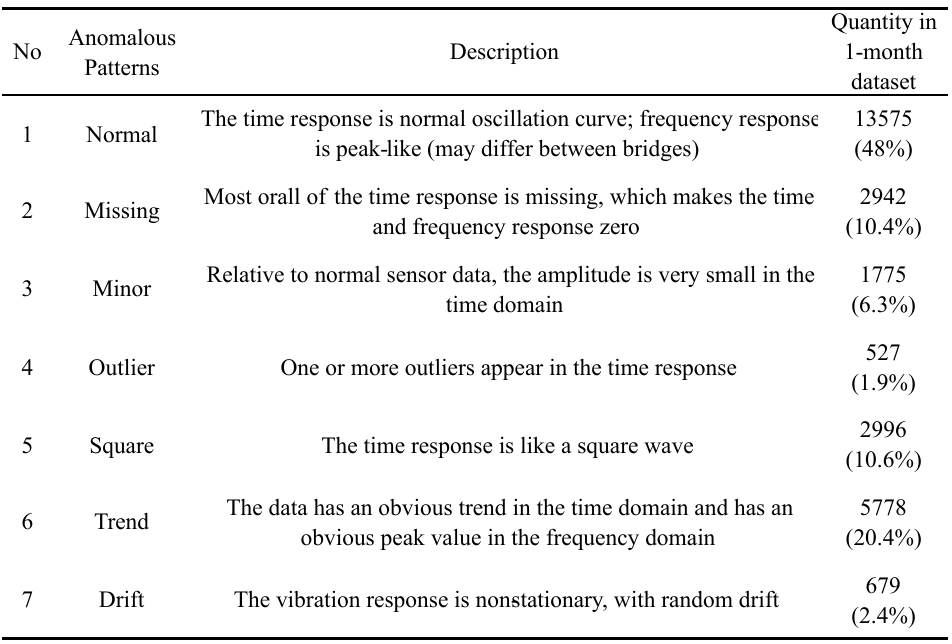}
\end{center}
\end{table}

\begin{figure}[htbp]
  \centering
  \captionsetup{justification=centering}
  \includegraphics[scale=0.8]{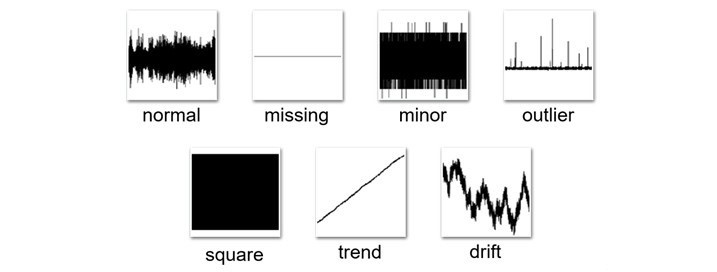}
  \caption{Examples for each of the anomaly pattern in the SHM data}
  \label{fig:fig3}
\end{figure}

\section{Methodology for anomaly detection}

The methodology for anomaly detection in SHM data involves 3 major steps as shown in Fig. 4. In the first step, the raw time series measurements are broken down into 1-hour segments as mentioned before. The peak envelopes of the time series are extracted to easily visualize the shape of the time series. 
\begin{figure}[htbp]
  \centering
  \captionsetup{justification=centering}
  \includegraphics[width=\textwidth,scale=0.9]{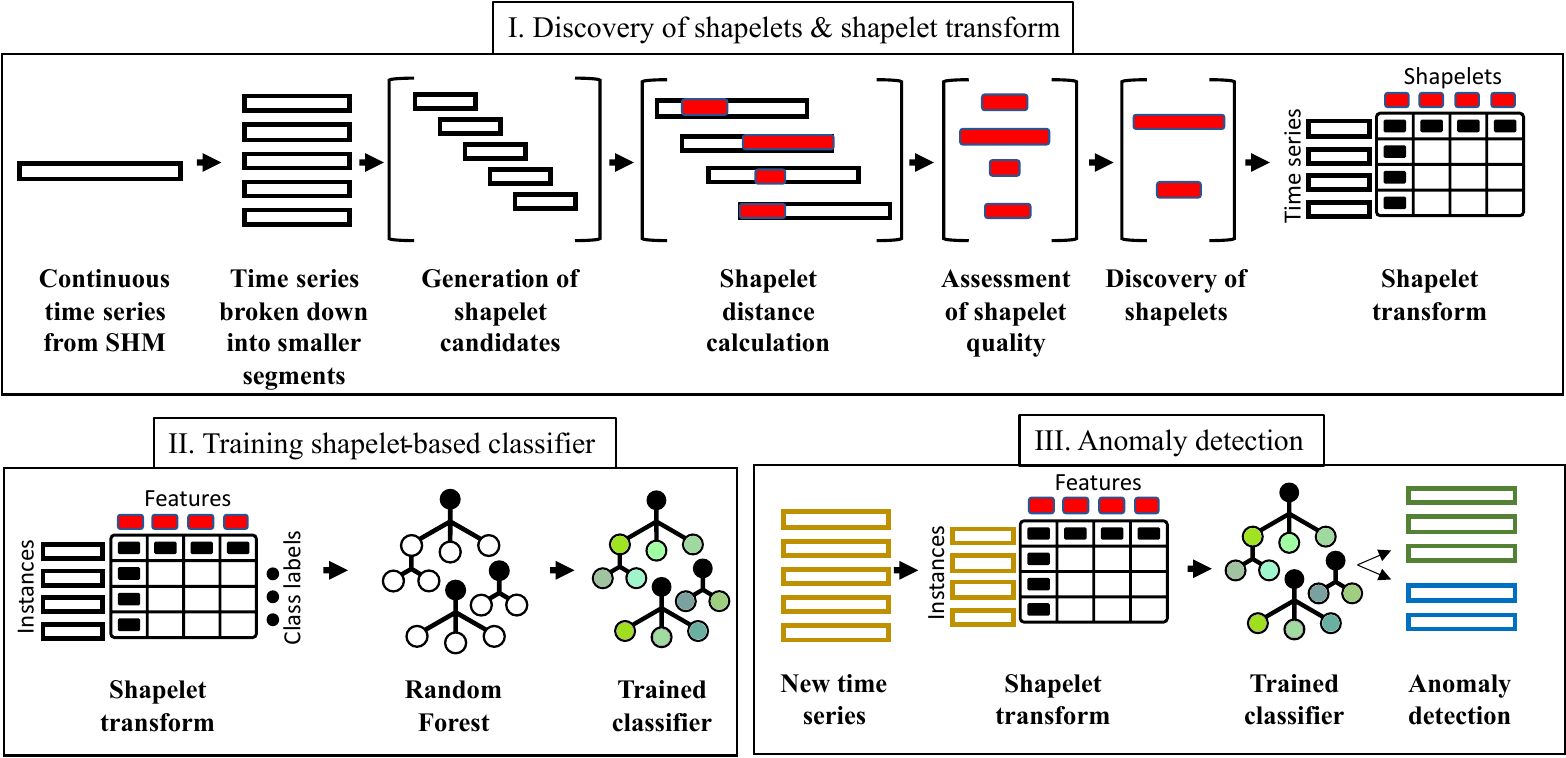}
  \caption{Methodology for anomaly detection in SHM data}
  \label{fig:fig4}
\end{figure}
A time series learning set is constructed along with class labels using these envelopes as shapes. Once the learning set is ready, the process of transforming it into local-shape space begins. Shapelet transform has five major stages: generation of shapelet candidates, distance calculation between a shapelet and a time series, assessment of the quality of shapelets, discovery of shapelets, and data transformation. In the second step, the original time series-based learning set is transformed into a local-shape space where each element is the distance between a shapelet and a time series. In this transformed learning set, the features are the discovered shapelets and the instances are the individual time series envelopes. This is fed to a Random Forest classifier for training. Once the training is complete, the trained classifier is used to classify normal and anomalous data from the new incoming time series from the SHM system in the third step.

\subsection{Step 1: Discovery of shapelets and shapelet transform}

\subsubsection{Preprocessing of raw data}

Based on visual inspection of Fig.3, it is easy to differentiate between the different anomalies. However, it is quite difficult to use the raw time histories for shapelet detection due to the long periods of periodic waveforms present in the vibrations. This can be overcome by extracting the envelopes of the acceleration time history which gives an overall shape to the vibration time series. The envelopes can then be easily used as input for the discovery of shapelets. Fig. 5 shows the extraction of peak envelopes of the bridge acceleration time history calculated using a moving window. Peak envelope is used here instead of a root-mean-square (RMS) envelope, as peak provides better differentiation between anomalous patterns when noisy or spurious signals are present. By looking at the peak envelopes of anomalies, the classification of anomalous data has become a much easier task now. Considering the computational demand of the algorithm, the envelopes obtained from the raw time series is down sampled to 1 Hz to improve the efficiency of the algorithm. Down sampling the data did not affect the shapes of the envelopes and hence the reliability of the method remains unchanged.

\begin{figure}[htbp]
  \centering
  \captionsetup{justification=centering}
  \includegraphics[scale=0.8]{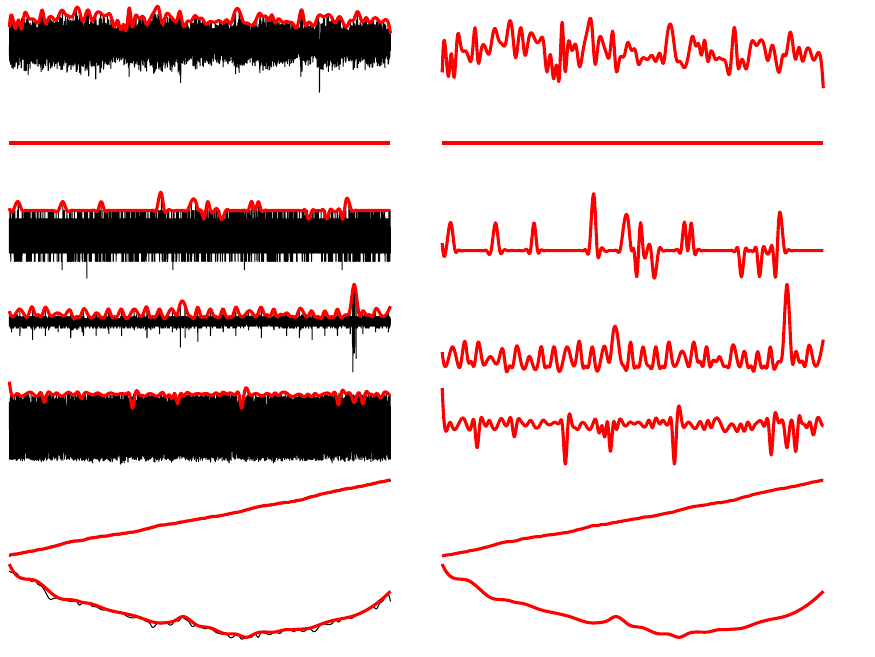}
  \caption{Extraction of peak envelopes from anomalous patterns}
  \label{fig:fig5}
\end{figure}

\subsubsection{Generation of shapelet candidates}

 Consider a time-series dataset ${TS}$. Let \textit{C} be the set of corresponding class labels for each time series. A time series learning set $\Phi \left\{ TS,C \right\}$ is first created by a vector of instance input-output pairs ${{\Phi }_{i}}=({{TS}_{i}},{{C}_{i}})$. Each subsequence in each time series in ${\Phi}$ is considered as a potential shapelet candidate. So, there are $\left( m-l \right)+1$ discrete subsequences of length \textit{l} between a subsequence \textit{X} of length \textit{l} of a time series \textit{TS} of length \textit{m}. If ${{W}_{1}}$ is the set of all candidate shapelets of length \textit{l} in a time series ${{TS}_{1}}$, then 

\begin{equation}
	\ {W}_{1}=\left\{ {{w}_{\min }},{{w}_{\min +1}},...,{{w}_{\max }} \right\}\
\end{equation}

where $min\ge 3$  as it is the minimum meaningful length for a time series and  $max\le m$.r

It should be noted that the shapelet algorithm independently normalizes all candidate shapelets so as to be invariant to scale and offset. For the present case, the time series learning set consists of 700 labeled set of time series envelopes that are extracted from the raw measurements as shown in Fig. 6. It should be noted that the learning set contains equal samples of patterns obtained from the SHM data i.e., the set contains 100 samples of normal pattern, 100 samples of missing pattern, 100 samples of minor pattern and so on. This is done to achieve a balanced training set to avoid classifier bias during the detection of anomalies. The reason for choosing 100 as the sample number is as follows. The data from 2012-01-01 to 2012-01-16 is used for training the algorithm and the data from the other fifteen days (2012-01-17 to 2012-01-31) is used for testing. In the training dataset, the “outlier” pattern had lowest quantity of about 100 datasets. Hence this number has been established as a baseline for choosing the number of samples for each pattern.  

Thus the time-series learning set, ${\Phi}$ has a total of 700 datasets as shown in Fig.6. Each time series in the training set has 3600 data points as the sampling frequency is 1 Hz. Let us take the first time series in the training set for illustration. As per Eq. (2) 

\begin{equation}
	\ {W}_{1}=\left\{ {{w}_{3}},{{w}_{4}},...,{{w}_{3599}},{{w}_{3600}} \right\}\
\end{equation}

where ${{w}_{3}}$(first 3 data points) is the shortest shapelet length and ${{w}_{3600}}$ (entire time series) is the longest shapelet length. Thus, the set ${{W}_{1}}$ contains 3598 different lengths of shapelets obtained from the first time series. In a similar way,shapelet candidates are generated from all the time series in the learning set.

\begin{figure}[htbp]
  \centering
  \captionsetup{justification=centering}
  \includegraphics[scale=0.5]{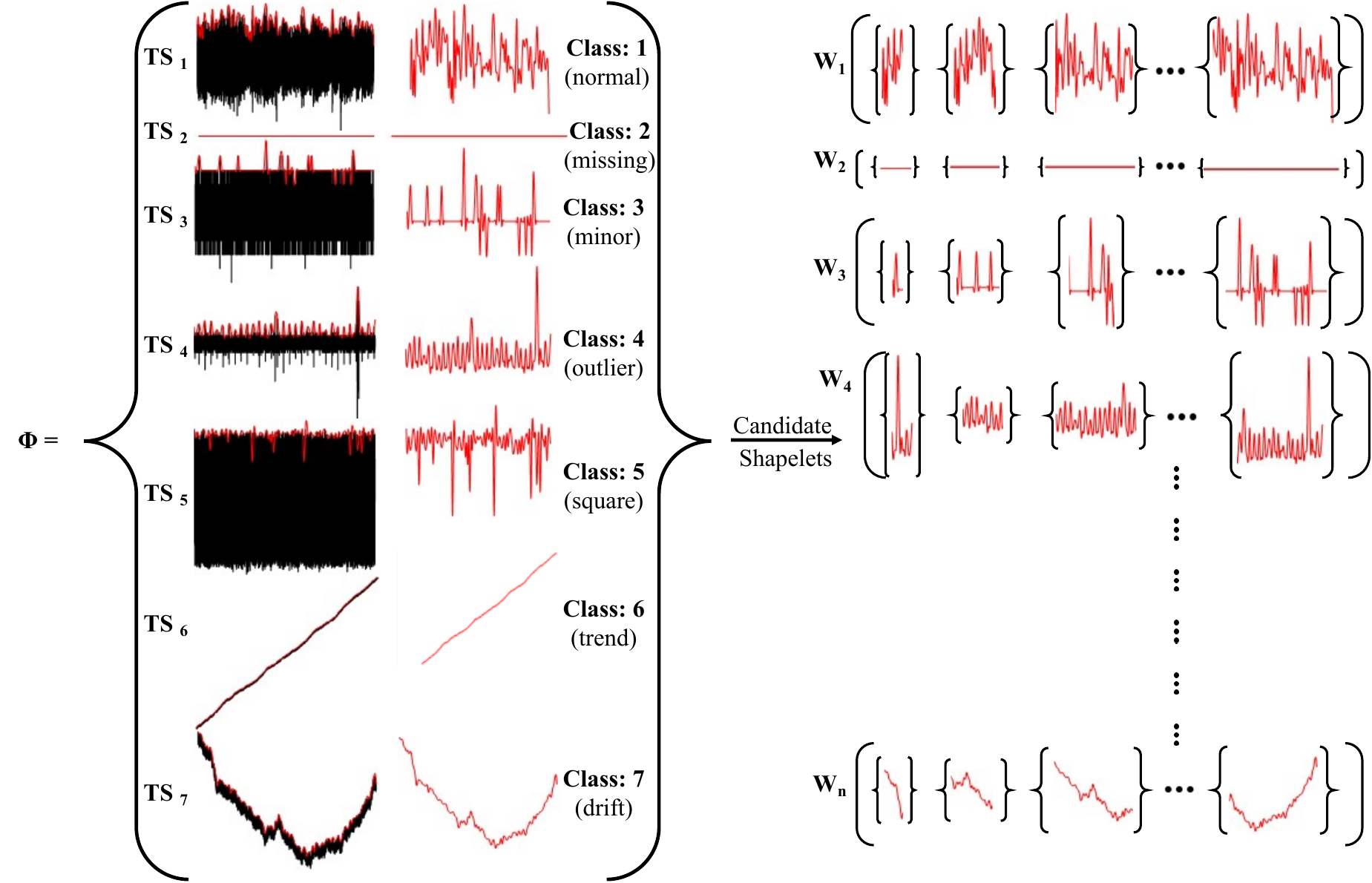}
  \caption{Illustration of generation of shapelet candidates for each time series in the time series learning set}
  \label{fig:fig6}
\end{figure}

\subsubsection{Shapelet distance calculation}

Euclidean distance is used as a similarity measure in shapelets and the squared Euclidean distance between a subsequence \textit{X} of length \textit{l} and another subsequence \textit{Y} of the same length is defined as:

\begin{equation}
	\ d(X,Y)=\sum\limits_{i=1}^{l}{{{\left( {{x}_{i}}-{{y}_{i}} \right)}^{2}}}\
\end{equation}

\begin{figure}[htbp]
  \centering
  \captionsetup{justification=centering}
  \includegraphics[scale=0.65]{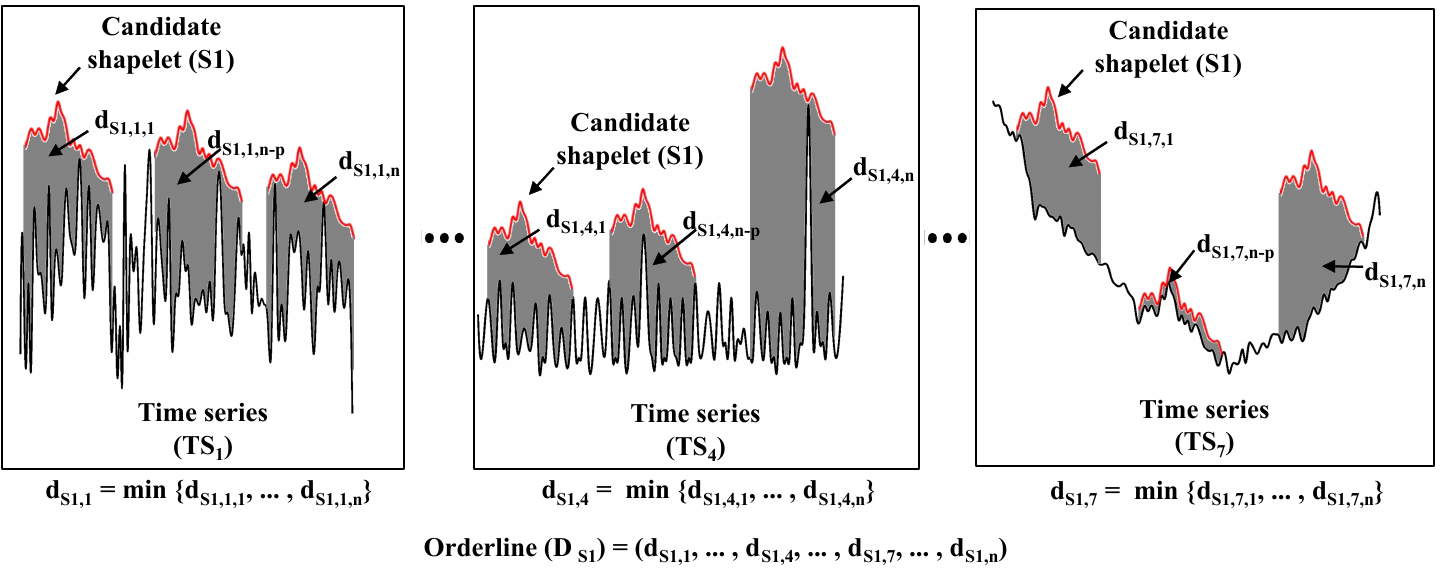}
  \caption{Illustration of Euclidean distance calculation between a candidate shapelet S1 and time series in the learning set}
  \label{fig:fig7}
\end{figure}

The distance between a potential shapelet candidate and all normalized series in \textit{TS} is computed to create a list of \textit{n} distances called an orderline ${D}_{S}$. An orderline consists of distance values and the class label corresponding to the time series for which the distance value is calculated. The orderline is then sorted in increasing order of the distance value. In the present study, each time series leads to the generation of 3598 shapelet candidates. Each of these 3598 shapelets is then compared with other time series using Euclidean distance. For illustration purpose, consider a shapelet candidate (${{S}_{1}}$) as shown in Fig. 7. The shapelet candidate moves over every time series and the minimum distance between the candidate and all the normalized lengths of subsequences in the time series set is calculated. It should be noted that Fig.7 is an exaggerated illustration to facilitate easy understanding and not an accurate depiction of the normalization process. If the shapelet candidate is generated from a pattern that is different from the time series being compared to, then it will lead to a large Euclidean distance. However, if the shapelet is similar to the one being compared to, then it will have a minimum Euclidean distance as seen in ${{d}_{S1,7}}$ in Fig. 7. Thus, the distance between a shapelet candidate ${{S}_{1}}$ and all the time series in \textit{TS} is given by,

\begin{equation}
    	\ {{D}_{S}}=\left\langle {{d}_{S1,1}},{{d}_{S1,2}},...,{{d}_{S1,n}} \right\rangle \
\end{equation}

It is a time-consuming task to calculate ${D}_{S}$ and hence a number of speed-up techniques have been proposed in the literature to handle the large volume of calculations. \cite{ye2009time,mueen2011logical,ye2011time,hills2014classification,rakthanmanon2013fast}

\subsubsection{Assessment of shapelet quality}

Information Gain (IG) \cite{shannon1949mathematical} is used as the standard approach to calculate the quality of a shapelet \cite{ye2009time,mueen2011logical,ye2011time}. If a time series dataset \textit{T} can be split into two classes, \textit{1} and \textit{2}, then the entropy of \textit{T} is: 

\begin{equation}
    \ H(T)=-p(1)\log (p(1))-p(2)\log (p(2))\
\end{equation}
	
where \textit{p(1)} and \textit{p(2)} are the proportion of objects in class \textit{1} and \textit{2} respectively. Thus every splitting strategy partitions the dataset \textit{T} into two sub-datasets ${T}_{I}$ and ${T}_{II}$. The Information Gain of this split is the difference between the entropy of the entire dataset and the sum of the weighted average of entropies for each split. In the present case, the splitting rule is based on the distance from the shapelet candidate ${S}$ to every series in the dataset.  The best possible shapelet will generate small distance values when compared to a time series of its own class and large distance values for time series from the other class. Thus the best arrangement for the orderline is to have all the distance values corresponding to the class of the shapelet in ${T}_{I}$ and the other in ${T}_{II}$. Thus, the information gain for each split is calculated as:

\begin{equation}
    \ IG=H(T)-\left( \frac{|{{T}_{I}}|}{|T|}H({{T}_{I}})+\frac{|{{T}_{II}}|}{|T|}H({{T}_{II}}) \right)\
\end{equation}

where $0\le IG\le 1$.

\vspace{\baselineskip}

\begin{figure}[htbp]
  \centering
  \captionsetup{justification=centering}
  \includegraphics[scale=0.9]{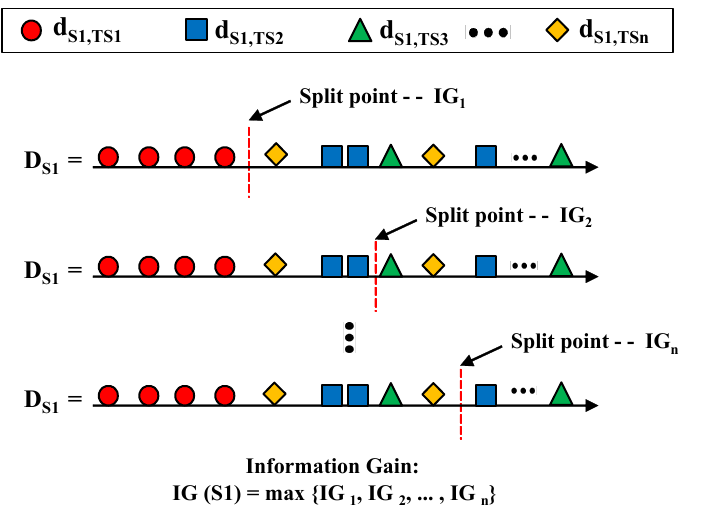}
  \caption{One-dimensional representation of the arrangement of time series objects by the distance to the candidate shapelet. Information Gain is calculated for each possible split point}
  \label{fig:fig8}
\end{figure}

The same procedure is extended to the 7-class problem as in the present study. For illustration purpose, consider the shapelet candidate (${S}_{1}$), mentioned in the previous section. ${S}_{1}$ is compared with 699 other time series in the learning set and thus 699 distances are obtained. These distances values are ordered in increasing value in the orderline and the information gain is calculated as shown in Fig.8. The same procedure is extended to all the shapelets candidates that is generated. Whichever length of shapelet surpasses the provided information gain threshold (0.05 in the present case) is retained and the other shapelet lengths are discarded. This makes sure that the selected shapelets are meaningful and have discriminatory power. Predetermining the optimal length of shapelet is impossible and unnecessary as it hinders the detection accuracy of the algorithm. It is also very difficult to interpret the variety of shapelet lengths obtained from the algorithm as these lengths have been chosen from several 1000s of shapelet lengths that were compared with several other time series. However, there is a provision in the shapelet algorithm to set the maximum and minimum shapelet length to achieve speedup. This provision should be used with care and should only be utilized in cases where only a certain length of shapelets are of interest.

\subsubsection{Discovery of shapelets and shapelet transform}

\begin{figure}[htbp]
  \centering
  \captionsetup{justification=centering}
  \includegraphics[scale=1.15]{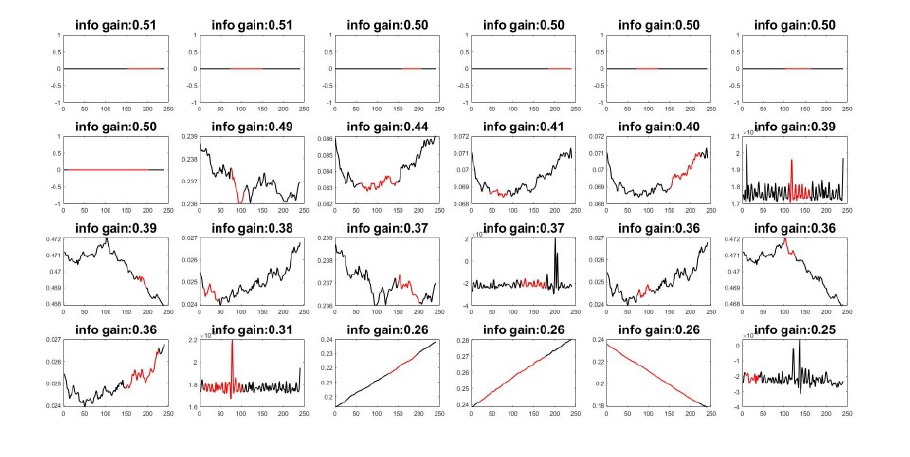}
  \caption{Shapelets (1-24) discovered for anomaly detection}
  \label{fig:fig9}
\end{figure}

\begin{figure}[htbp]
  \centering
  \captionsetup{justification=centering}
  \includegraphics[scale=1.15]{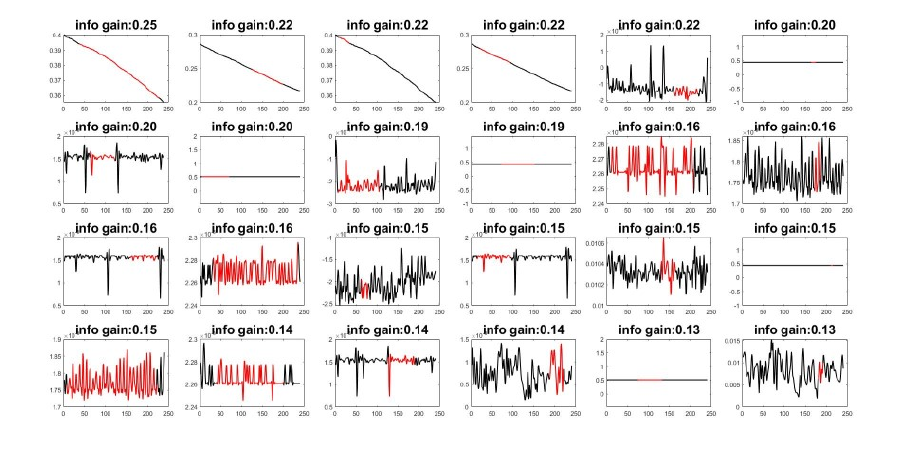}
  \caption{Shapelets (25-48) discovered for anomaly detection}
  \label{fig:fig10}
\end{figure}

\begin{figure}[htbp]
  \centering
  \captionsetup{justification=centering}
  \includegraphics[scale=1.15]{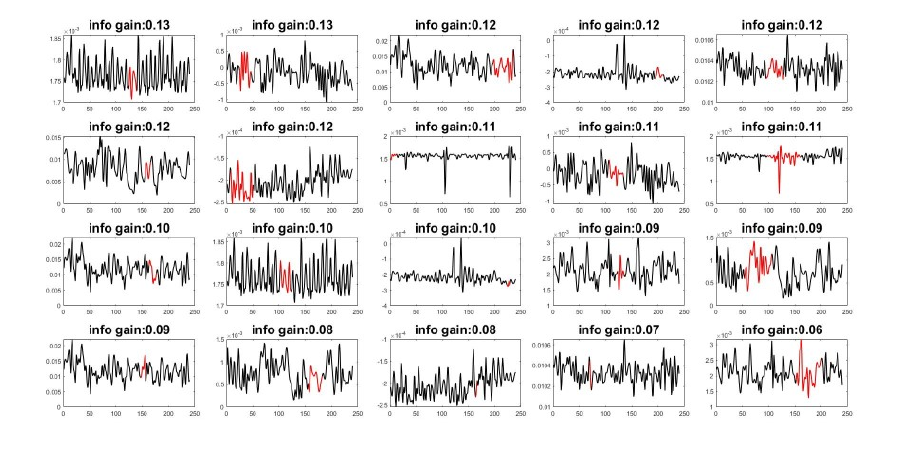}
  \caption{Shapelets (49-68) discovered for anomaly detection}
  \label{fig:fig11}
\end{figure}

 An algorithm combining all of the above mentioned components of shapelet discovery was developed by \cite{bostrom2017binary,lines2012shapelet,hills2014classification} and is available at \url{www.timeseriesclassification.com}. The same algorithm has been adopted and modified to suit the datasets under consideration for the present study. The same algorithm has been adopted and modified to suit the datasets under consideration for the present study. The input to the algorithm is the time series leaning set ${\Phi}$. As mentioned in the previous sections, the default minimum length of the shapelets is set to 3 and the maximum length is equal to the length of each individual time series. The number of shapelets to store (\textit{r}) is set to a default of 10 times the number of time series in the training set. Moreover, based on the number of classes (\textit{numC}) in the training set, a limit of \textit{r/numC} shapelets for each class is set as the maximum number of shapelets to store per class. The minimum information gain threshold is set to a default value of 0.05. This makes sure that poor quality shapelets below this threshold are removed during the shapelet finding process. Using the provided parameters, the algorithm then makes a single pass through the time series data in ${\Phi}$ taking each subsequence of every time series as a potential shapelet candidate. The generated shapelet candidates are also normalized to make them independent of scale and offset. The distance between each shapelet candidate and time series in the training dataset is calculated and the order list ${D}_{S}$ is formed to assess the quality of shapelets using Information Gain. Once all the shapelets in a time series have been assessed, the poor quality shapelets are removed and the rest is added to the shapelet set. After all the time series in the training set have been evaluated this way, the algorithm returns the discovered shapelets. .

For the present study, the shapelet algorithm is implemented in Python as a single-core serial job on an Intel Xeon Processor E5-2620 (2.6-GHz CPU) for 1 hour, and the algorithm discovered a total of 68 shapelets. Various shapes were discovered for each of the seven anomalous patterns, as shown in Figs. 9 – 11. Shapelets corresponding to the “missing” and “normal” patterns have the highest information gain as these shapes separate the classes easily. 

 \begin{figure}[htbp]
  \centering
  \captionsetup{justification=centering}
  \includegraphics[scale=0.9]{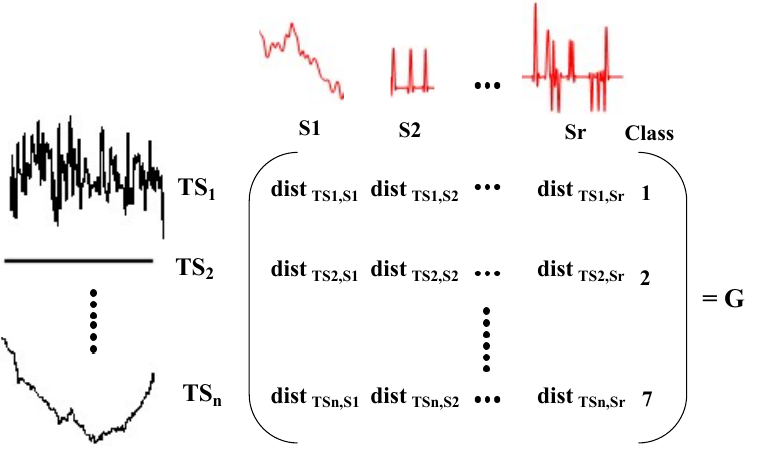}
  \caption{Shapelet Transform containing a matrix of Euclidean distance between the discovered shapelets and the other time series in the learning set}
  \label{fig:fig12}
\end{figure}

 Once the shapelets are discovered, the next step is to transform the learning set ${\Phi}$ into a local-shape space where each feature is the distance between a shapelet and a time series. So, given a set of a time series dataset \textit{TS} containing \textit{n} time series and a set of \textit{r} discovered shapelets, the shapelet transform algorithm calculates the minimum distance between each discovered shapelet and each time series in the dataset. This transformation creates a matrix \textit{G} that contains \textit{n} rows and \textit{r} columns matrix as illustrated in Fig. 12 where each element is the minimum Euclidean distance between each shapelet and time series, with the class values appended to the end of each row. It should be noted that ${S}_{1}$ is the first shapelet in the set of all shapelets and does not denote that it is obtained from ${TS}_{1}$. The matrix \textit{G} now serves as the standard instance-attribute dataset that is used in machine learning tasks that can be used with any supervised or unsupervised algorithm. In the present study, shapelet transform constructs a 3500 x 68 matrix where each element corresponds to the minimum Euclidean distance between each shapelet and the time series.

\subsection{Step 2: Training of shapelet-based classifier}

 The shapelet based classifier originally developed by \cite{ye2009time} embeds shapelet finding in a decision tree classifier where shapelets are found at every node. Many researchers ever since have demonstrated that higher accuracy can be achieved by using shapelets with more complex classifiers or ensemble of classifiers than with decision trees, where over fitting is a major issue.  \citep{lines2012shapelet,hills2014classification,bagnall2017great,bostrom2017shapelet}. For the present study, Random Forest \citep{breiman2001random} is used as a classifier for time series classification. The Random Forest algorithm seeks to solve the issues with decision trees by classifying examples through using a multitude of decision trees and predicting the class of a sample based on the mean probability estimate across all the trees. Thus, a Random Forest classifier with 500 trees is used for training on the shapelet-transformed dataset. \cite{hills2014classification,bagnall2017great} compared the performance of shapelet transform using several standard classifiers and ensemble classifiers on a variety of datasets from UCR time-series repository. According to their study, a shapelet-based random forest classifier with 500 trees is found to be optimal. Hence the same has been adopted in the present study. It is also found that increasing the number of trees beyond 500 did not result in any significant increase in accuracy.

\subsection{Step 3: Anomaly detection}

As mentioned in section 4.1.1, the data from 2012-01-01 to 2012-01-16 is used for training the algorithm and the data from the other fifteen days (2012-01-17 to 2012-01-31) is used for testing. The raw measurements are broken down into 1-hr segments resulting in a total of 13,679 datasets. The peak envelopes of the time series are extracted and down sampled to 1 Hz. The shapelet transform algorithm is used on the test set to transform the data onto shape-space where each element is the minimum Euclidean distance between the discovered shapelets and the time series in the test set. Thus a 13,679 x 68 matrix is obtained where 13,679 are the time series instances and 68 are the shapelet-based features. The trained Random Forest classifier is then tested on this transformed test set.  

\section{Results and discussion}

The detection of anomalies using shapelet-based classifier was implemented as a single-core serial job on an Intel Xeon Processor E5-2620 (2.6-GHz CPU) and the algorithm took 2.5 hours to output the results. The run time will drastically increase if event detections are made for months or years of continuous data. The run time can be reduced in two ways. Incorporating parallelism in the algorithm so that distance calculations can be executed in parallel on multicore machines. Another way is to redesign the algorithm to make it suitable for parallel Graphics Process Units (GPUs). \cite{chang2012efficient} improved the shapelet algorithm for GPU implementation and achieved speedups nearly two orders of magnitude faster than CPU implementation. This means that a 1.7-hour CPU implementation of shapelets will only take 2 minutes using GPUs. Such an algorithm redesign will be explored in future studies to render this method efficient for processing large volumes of data.

The detection results are shown in Table 2. The definitions in the following section will help understand the performance metrics of the classifier better.

\begin{table}[htbp]
\begin{center}
\captionof{table}{Performance metrics of the shapelet-based Random Forest classifier} \label{resbar}
\includegraphics[scale=0.9]{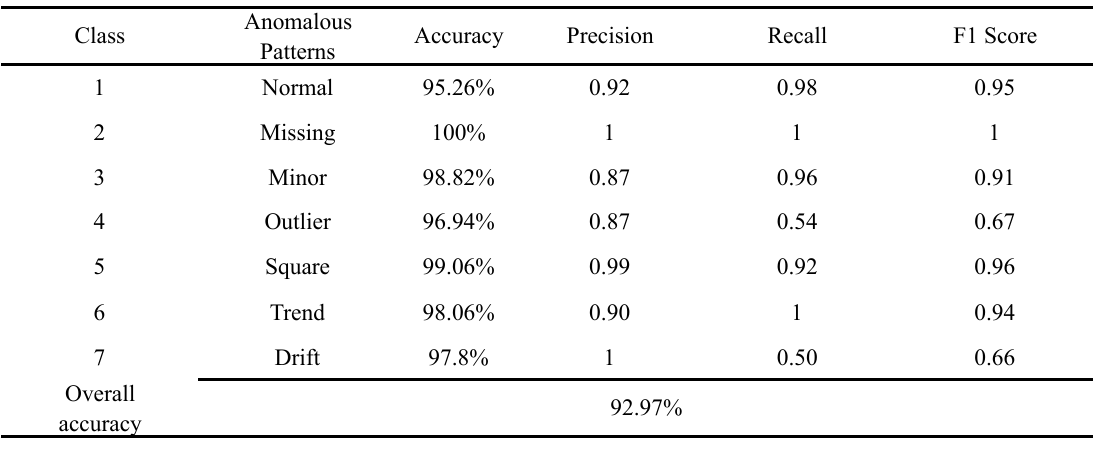}
\end{center}
\end{table}

\subsection{Terminologies used in assessing the performance of the classifier}
\subsubsection{True Negative (TN)}
The actual value is False, and the classifier also predicted False.

\subsubsection{False Positive (FP)}
 The actual value is False, and the classifier predicted True.
 
 \subsubsection{False Negative (FN)}
 The actual value is True, and the classifier predicted False.
 
 \subsubsection{True Positive (FP)}
 The actual value is True, and the classifier also predicted True.

\subsubsection{Accuracy}

Accuracy is the sum true positives and true negatives divided by the total number of instances. From the confusion matrix, accuracy is the sum of the elements on the diagonal divided by the total number of predictions made. Accuracy is calculated as follows.
\begin{equation}
    \ Accuracy=\frac{TP+TN}{TP+TN+FP+FN}\
\end{equation}

\subsubsection{Precision}

Precision is the ratio of number of correct predictions to the number of total predictions made. If a class has high precision, then it means that if the classifier predict this class, it is most likely to be true. Precision is given by:
\begin{equation}
    \ Precision=\frac{TP}{TP+FP}\
\end{equation}

\subsubsection{Recall(Sensitivity)}

Recall is the ratio of correctly predicted positive observations to all actual class observations. It is the measure of how many true positives get predicted from all the positives in the dataset. Recall will have a value of 1 for an ideal classifier with no false negatives. 
\begin{equation}
    \ Recall=\frac{TP}{TP+FN}\
\end{equation}

\subsubsection{F1 Score}

F1 score is the harmonic mean of precision and recall and is a combined measure of the two. F1 score is high when both precision and recall are high. 
\begin{equation}
    \ F1 Score=2*\frac{Precision*Recall}{Precision+Recall}\
\end{equation}

\subsection{Discussion of results}
\begin{figure}[htbp]
  \centering
  \captionsetup{justification=centering}
  \includegraphics[scale=0.9]{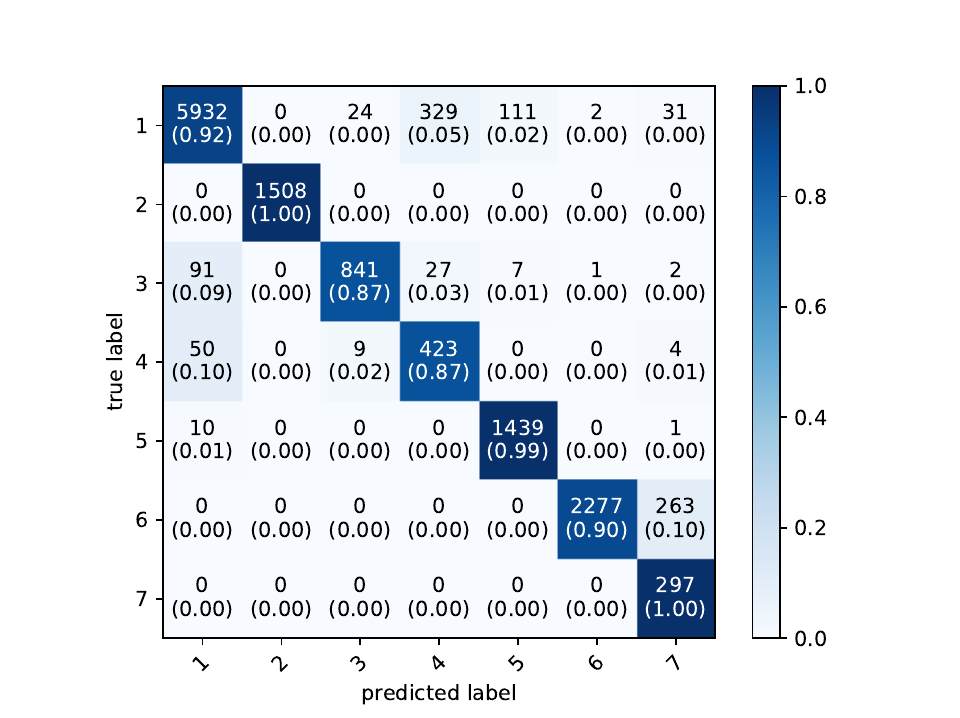}
  \caption{Confusion matrix for detected anomalies}
  \label{fig:fig13}
\end{figure}
The performance metrics are also shown visually in the form of a confusion matrix in Fig. 13. In the confusion matrix, the diagonal elements are the correctly classified instances and their corresponding precision is provided underneath within brackets. An overall accuracy of 93\% is obtained using the shapelet-based classifier. The individual accuracies of all the classes are above 95\% with class 2 and 5 having an accuracy of about 100\%. In terms of precision and recall, classes “normal”, “square” and “trend” have a high value of over 90\% with class “missing” having a maximum of 100\%. For classes “outlier” and “drift”, even though the precision is high, recall value is very low. This is due to the fact that, a small number of instances in class “normal” were predicted as belonging to class “outlier” due to the presence of significant outliers. Also, some of instances in class “normal” were predicted as “class square” as the time series has a very close resemblance to a square shape. Similarly, a number of instances in class “trend” were predicted as belonging to class “drift” as the time series closely resembled class “drift”. Each of these cases are examined in detail and remedial measures are proposed in the following sections. Meanwhile, in the present study, since the learning set is constructed as a well-balanced dataset, of all the performance metrics, accuracy measure can be used as a useful indicator to comment on the performance of the classifier. Based on high individual and overall accuracy, the proposed shapelet-based classifier has a very good ability to identify anomalies in SHM data. 

\section{Remedial measures for increasing the performance of the classifier}
\subsection{Removal of outliers during pre-processing}

From the confusion matrix, it can be seen that 329 instances in class “normal” are predicted as class “outlier”. On closer inspection, it is found that the outliers not only affect the instances in class “outlier” but also the instances in class “normal”. One such example of a class “normal” instance with outliers is shown in Fig. 12 in the upper left corner. This confuses the machine learning algorithm as it has learned that “outlier” is the only class with large outliers. Hence it is wise to remove all the predominant outliers in the preprocessing step so that class “outlier” becomes a pure class that only contains datasets with significant outliers. This can be easily done using the ‘rmoutliers’ command in Matlab that detects and removes predominant outliers according to a user specified window. It should be noted that this command not only removes outliers from class “normal, it also removes significant outliers in class “outlier”. From Fig. 14, it can be seen that in the first column, after removal of outliers, class “normal” appears clean. This will increase the accuracy of the classifier as it will not be confused over the presence of outliers in class “1”. In the second column, a single outlier in an instance in class “outlier” is removed which transforms the time series to class “minor”. In the third column, even after the removal of predominant outliers, certain pesky outliers remain and hence this instance still belongs to class “outlier”.  Relabeling datasets in this way, after outlier removal, will lead to pure classes which in turn leads to better classifier performance. 

\begin{figure}[htbp]
  \centering
  \captionsetup{justification=centering}
  \includegraphics[scale=0.6]{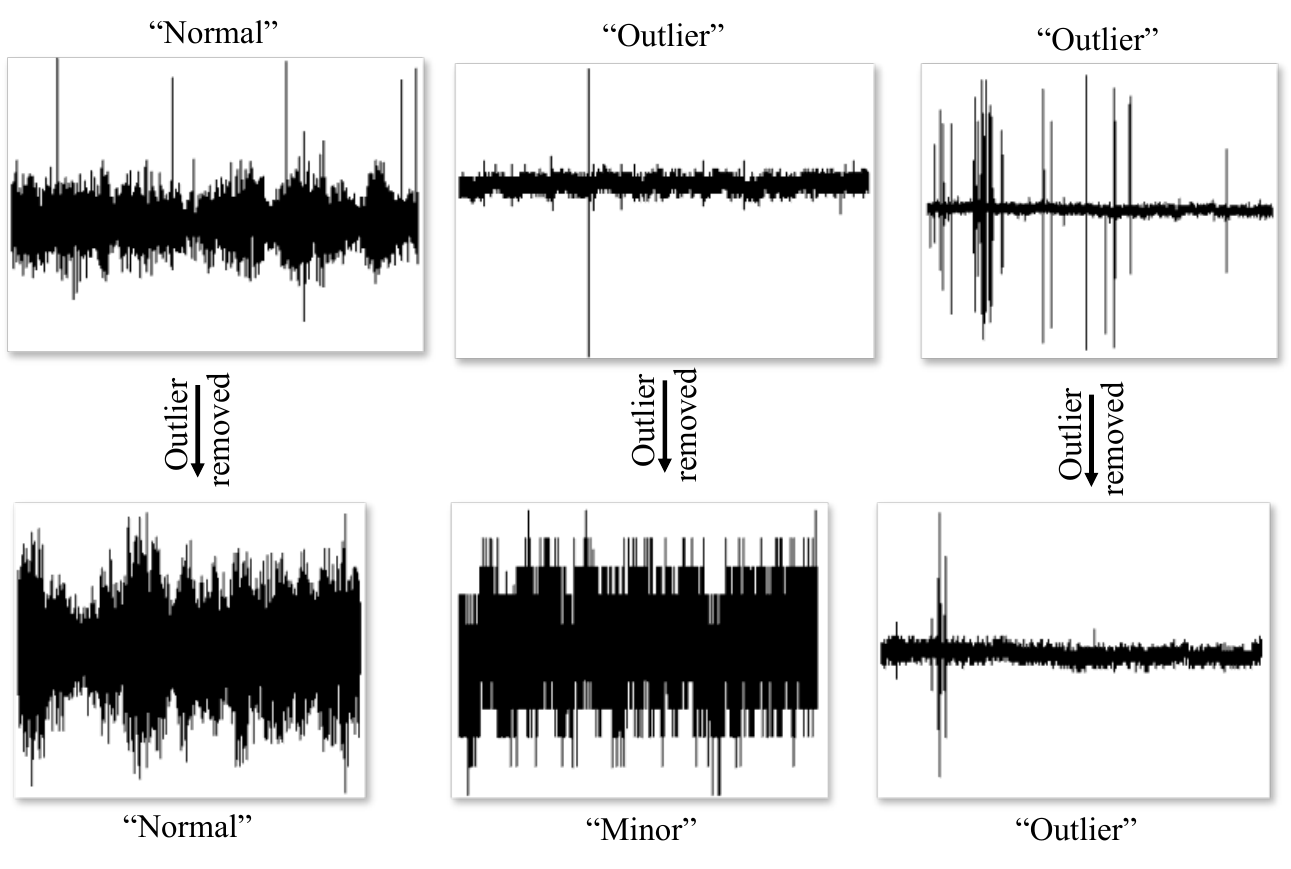}
  \caption{Removal of outliers during pre-processing}
  \label{fig:fig14}
\end{figure}

\subsection{Detrending time series during pre-processing}

It can be seen from the confusion matrix that 263 instances in class “trend” are labelled as “class “drift”. On close inspection, it is noted that the instances in class “trend” are nothing but the instances in “drift”  with a trend in time series. Since the algorithm is trained on time series envelopes, the classifier finds many similarities between the two classes. Moreover, the class “trend” contains varieties of time series trends increasing from left to right and vice versa. This introduces difficulty during the learning process. In Fig. 15, a time series instance in class “trend” is detrended and it transforms to class “drift”. After thorough inspection, it is found that this is the case for all of the instances in class “trend”. So, if all the time series instances are detrended this way, during the pre-processing step, this will transform the 7-class problem to a 6-class problem. This is a huge advantage in terms of computational efficiency and classifier performance. Detrending can be applied together with removal of outliers in the pre-processing step and datasets need to be relabeled before feeding it to the machine learning algorithm. These simple preprocessing steps will drastically increase the performance of the classifier. 

\begin{figure}[htbp]
  \centering
  \captionsetup{justification=centering}
  \includegraphics[scale=0.6]{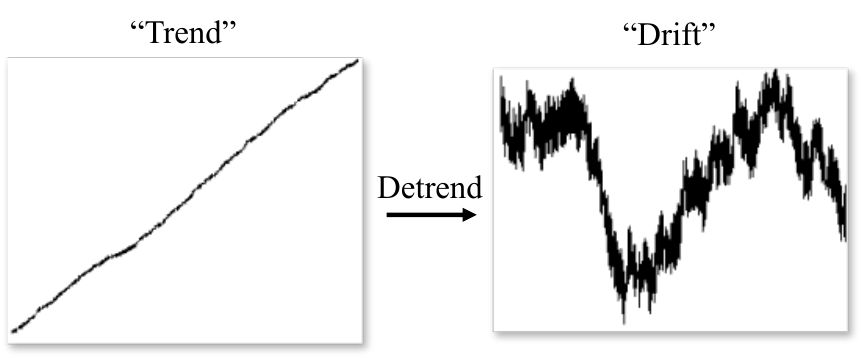}
  \caption{Detrending during pre-processing}
  \label{fig:fig15}
\end{figure}

\section{Conclusion}
Anomaly detection is a long standing problem in the SHM community. In this paper, this fundamental problem is addressed by autonomously identifying anomalous data patterns in 1-month of acceleration data from a SHM system installed on a long-span bridge in China. This is achieved using a relatively new and efficient time series representation named “Shapelet transform” that is combined with machine learning algorithm (Random Forest classifier) to identify anomalies in SHM data. Shapelet transform is a unique time series representation technique that is solely based on the shape of the time series and provides a universal standard feature for detection which is based on the distance between a shapelet and a time series. The raw measurements of every sensor anomaly has a unique time series shape and the shapelet transform utilizes this feature to easily capture these distinct shapes. These shapes are used to transform the SHM data into a local-shape space and the Random Forest classifier is then trained on this transformed dataset to identify and classify the different anomalous data patterns.

 The data used in the current study has 6 different anomalous patterns of time series. From the 1-month acceleration data, the first sixteen days is used for training the algorithm and the data from the other fifteen days is used for testing. A balanced dataset is created that contains equal samples from all classes of anomalies. The shapelet algorithm discovered 68 shapes from the training dataset. These shapes are used to transform the dataset into a local shape-space. The transformed dataset is then used to train a Random Forest classifier for anomaly detection. The classifier has an overall accuracy of 93\% which indicates that the proposed shapelet-based classifier has a very good ability to identify anomalies in SHM data. The individual accuracy of all the classes are also above 95\%. Various pre-processing measures are also proposed in this paper to increase the classifier performance even further and this will be pursued in future studies.
 
\section*{Data and resources}
The structural health monitoring data of the long-span bridge is obtained from the organizers of the 1st International Project Competition for Structural Health Monitoring (IPC - SHM), 2020 (\url{http://www.schm.org.cn/#/IPC-SHM,2020}). The basic algorithm for shapelet discovery is available at \url{www.sktime.org}. Additional information related to this paper may be requested from the authors.

\section*{Acknowledgments}
This work was supported in part by the Robert M. Moran Professorship and National Science Foundation Grant (CMMI 1612843). The authors would like to thank the organizers of the International Project Competition for Structural Health Monitoring (IPC-SHM 2020), Asia-Pacific Network of Centers for Research in Smart Structures Technology (ANCRiSST), Harbin Institute of Technology (China), and the University of Illinois at Urbana-Champaign (USA) for providing the structural health monitoring data of the long-span bridge. The authors also would like to thank the chairs of IPC-SHM 2020, Prof. Hui Li and Prof. Billie F. Spencer Jr, for their leadership in the competition.

\bibliographystyle{apalike}  
\bibliography{references}  %%% Remove comment to use the external .bib file (using bibtex).
%\setcitestyle{authoryear,open={(},close={)}}


% \bibitem{allen1978automatic}
% George Kour and Raid Saabne.
% \newblock Fast classification of handwritten on-line arabic characters.
% \newblock In {\em Soft Computing and Pattern Recognition (SoCPaR), 2014 6th
%   International Conference of}, pages 312--318. IEEE, 2014.

% \bibitem{hadash2018estimate}
% Guy Hadash, Einat Kermany, Boaz Carmeli, Ofer Lavi, George Kour, and Alon
% \newblock Estimate and replace: A novel approach to integrating deep neural
%   networks with existing applications.
% \newblock {\em arXiv preprint arXiv:1804.09028}, 2018.

% \end{thebibliography}

\end{document}